\documentclass[letterpaper]{article} 
\usepackage[draft]{aaai25}  
\usepackage{times}  
\usepackage{helvet}  
\usepackage{courier}  
\usepackage[hyphens]{url}  
\usepackage{graphicx} 
\usepackage{natbib}  
\usepackage{caption} 
\usepackage{algorithm}
\usepackage{algorithmic}

\usepackage{newfloat}
\usepackage{listings}
\DeclareCaptionStyle{ruled}{labelfont=normalfont,labelsep=colon,strut=off} 
\floatstyle{ruled}
\newfloat{listing}{tb}{lst}{}
\floatname{listing}{Listing}
\usepackage{graphicx} 
\usepackage{microtype}
\usepackage{url}
\usepackage{booktabs}
\usepackage{tabularx}
\usepackage{colortbl}
\usepackage{multirow}
\usepackage{xcolor}
\usepackage{bibentry}

\definecolor{lightbrown}{HTML}{cba487}
\definecolor{lightorange}{HTML}{f6c4a2}
\definecolor{lightyellow}{HTML}{fce597}
\definecolor{lightpurple}{HTML}{decbff}
\definecolor{lightgrey}{HTML}{d4d4d4}

\title{SafetyPrompts: a Systematic Review of Open Datasets for\\ Evaluating and Improving Large Language Model Safety}

\author{
    Paul Röttger\textsuperscript{\rm 1}, Fabio Pernisi\textsuperscript{\rm 1}, Bertie Vidgen\textsuperscript{\rm 2}, and Dirk Hovy\textsuperscript{\rm 1}
}
\affiliations{
    \textsuperscript{\rm 1}Bocconi University, 
    \textsuperscript{\rm 2}Contextual AI
}

\begin{document}

\maketitle

\begin{abstract}
The last two years have seen a rapid growth in concerns around the safety of large language models (LLMs).
Researchers and practitioners have met these concerns by creating an abundance of datasets for evaluating and improving LLM safety.
However, much of this work has happened in parallel, and with very different goals in mind, ranging from the mitigation of near-term risks around bias and toxic content generation to the assessment of longer-term catastrophic risk potential.
This makes it difficult for researchers and practitioners to find the most relevant datasets for their use case, and to identify gaps in dataset coverage that future work may fill.
To remedy these issues, we conduct a first systematic review of open datasets for evaluating and improving LLM safety.
We review 144 datasets, which we identified through an iterative and community-driven process over the course of several months.
We highlight patterns and trends, such as a trend towards fully synthetic datasets, as well as gaps in dataset coverage, such as a clear lack of non-English and naturalistic datasets.
We also examine how LLM safety datasets are used in practice -- in LLM release publications and popular LLM benchmarks -- finding that current evaluation practices are highly idiosyncratic and make use of only a small fraction of available datasets.
Our contributions are based on SafetyPrompts.com, a living catalogue of open datasets for LLM safety, which we plan to update continuously as the field of LLM safety develops.
\end{abstract}

%

\section{Introduction}

Ensuring the safety of large language models (LLMs) has become as a key priority for model developers and regulators.
Consequently, in recent years, researchers and practitioners have created an abundance of datasets for evaluating and improving LLM safety.
Safety, however, is a multi-faceted and contextual concept that lacks a unifying definition \citep{aroyo2023dices}.
This complexity is reflected in the current landscape of safety datasets, which is broad, diverse, and fast-changing.
Within just two months of 2024, for example, researchers published datasets for evaluating near-term risks from LLMs, such as sociodemographic bias \citep{gupta2024calm} and toxic content generation \citep{bianchi2024safetytuned}, as well as datasets for evaluating long-term societal risk potential, around power-seeking \citep{mazeika2024harmbench} and sycophantic behaviours \citep{sharma2024sycophancyeval}.
This rapid pace of dataset creation, and the variety of purposes served by different datasets, make it difficult for researchers and practitioners to find the most relevant datasets for different use cases, and to identify gaps in dataset coverage that future work may fill.

In this paper, we seek to address these issues by conducting a \textbf{first systematic review of open datasets for evaluating and improving LLM safety}.
We identify 144 datasets published between June 2018 and December 2024 based on clear inclusion criteria (\S\ref{subsec: methods - inclusion criteria}) using a comprehensive community-driven search method (\S\ref{subsec: methods - finding datasets}).
We examine these 144 datasets along several key dimensions, including their purpose (\S\ref{subsec: findings - purpose}), intended use (\S\ref{subsec: findings - intended use}), format and size (\S\ref{subsec: findings - format}), their creation (\S\ref{subsec: findings - creation}), language (\S\ref{subsec: findings - language}), licensing and access (\S\ref{subsec: findings - access}), and publication (\S\ref{subsec: findings - publication}).
Our review shows that a growing interest in LLM safety is driving the creation of more and more diverse open LLM safety datasets, with most contributions coming from academia.
Major outstanding challenges, on the other hand, include a clear lack of safety datasets in non-English languages as well as a lack of naturalistic safety evaluations.
We also review how open LLM safety datasets are used in practice -- in model release publications (\S\ref{sec: datasets in model releases}) and popular LLM benchmarks (\S\ref{sec: datasets in model benchmarks}) -- finding that current evaluation practices are highly idiosyncratic and leverage only a small fraction of available datasets.
We argue that this creates clear scope for standardisation in LLM safety evaluations, and that model safety evaluations in general could be improved by better leveraging recent progress in dataset creation (\S\ref{sec: discussion}).

\section{Dataset Review Methodology}
\label{sec: dataset review methods}

\subsection{Inclusion Criteria}
\label{subsec: methods - inclusion criteria}

At a high level, we restrict our review to \emph{open} datasets that are relevant to \emph{LLMs}, and specifically relevant to evaluating and improving LLM \emph{safety}.
More specifically, this means:

In terms of \textbf{data modality}, we only include text datasets.
We do not include image datasets \citep[e.g.][]{schwemmer2020diagnosing, zhao2021understanding, ricker2022deepfakes} or audio datasets \citep[e.g.][]{reimao2019dataset,koenecke2020racial,meyer2020artie}.
We also do not include datasets targeted at multimodal models, even if one modality is text, such as in the case of vision-language \citep[e.g.][]{carlini2023aligned, hall2023visogender,wolfe2023visionlanguagebias} or text-to-image models \citep[e.g.][]{bianchi2023texttoimage,parrish2023adversarial,luccioni2024stable}.
Further, we do not include datasets targeted at code generation models \citep[e.g.][]{siddiq2022securityeval, bhatt2023purple}.
These modalities and models constitute natural expansions for future work.

We make only minimal restrictions in terms of \textbf{data format}.
Real-world user interactions with LLMs usually take the form of text chat
\citep{ouyang2023shifted,zheng2024lmsys,zhao2024inthewildchat}, so we are most interested in datasets that naturally fit a chat format, like open-ended questions and instructions, but we also consider any other dataset that can meaningfully be expressed in a prompt format.
This includes multiple-choice questions or autocomplete-style text snippets.
We do not make any restrictions on dataset language.

For \textbf{data access}, we only include datasets that are publicly available for download via GitHub and/or Hugging Face.
We do not make restrictions based on how data is licensed.

Finally, we require that all datasets are \textbf{relevant to safety}.
For the purposes of our review, we adopt a wide and open definition of safety.
Broadly speaking, we include datasets that relate to representational, political or other forms of sociodemographic bias; to toxicity, malicious instructions or harmful advice; to hazardous behaviours like sycophancy or power-seeking; to alignment with social, moral or ethical values; or to adversarial LLM usage (e.g.\ red-teaming, jailbreaking, prompt hacking).
We only include datasets that explicitly focus on one or multiple of these aspects of LLM safety.
We do not include datasets that target general LLM capabilities like reasoning, language understanding, or code completion \citep[e.g.][]{dua2019drop,hendrycks2020mmlu,chen2021humaneval}.
We also do not include datasets that target factuality in LLMs, unless they directly relate to safety, like in the case of generating misinformation \citep[][]{souly2024strongreject} or measuring truthfulness \citep{lin2022truthfulqa}.%
\footnote{Each dataset candidate was reviewed by two authors of this paper.
We make detailed information on each dataset, as well as excluded datasets + exclusion reasons, available in the project repo.}

\textbf{The cutoff date for our review is December 17th, 2024}.
We only included datasets that were first published (i.e.\ made publicly available online) before this date.

\subsection{Finding Dataset Candidates}
\label{subsec: methods - finding datasets}

We used an iterative and community-driven approach combined with snowball search to identify dataset candidates for inclusion in our review.
In January 2024, we released a first version of SafetyPrompts.com, with an initial list of 44 datasets that we had compiled in a heuristic fashion based on prior work and our knowledge of the LLM safety field.
Over the next months, we marketed the website to the LLM safety community on Twitter and Reddit, to solicit feedback and further dataset suggestions.
This resulted in 51 additional datasets, suggested by many different researchers and practitioners.
We then used these 95 datasets as a starting point for snowball search, wherein we reviewed each publication corresponding to each dataset for references to other potentially relevant datasets.
Whenever we identified a new dataset, we repeated this process.
This resulted in 49 additional datasets.
\textbf{Overall, our review includes 144 open datasets for evaluating and improving LLM safety}, which were first published between June 2018 and December 2024.

We chose this review method because of two main reasons.
First, LLM safety is a very fast-moving field with contributions from across academia and industry.
By sharing intermittent results of our review on SafetyPrompts.com, we were able to solicit feedback from a broad range of stakeholders and expand our review, while also providing a useful resource to the community well ahead of the release of this paper.
Second, traditional systematic review methods like keyword search are ill-suited to the scope of our review.
Combinations of relevant keywords like ``language model'', ``safety'' and ``dataset'' return thousands of results on Google Scholar and similar platforms -- and still fail to capture the many types of datasets that may not mention ``safety'' but are highly relevant to it regardless, like toxic conversation datasets or bias evaluations.
Despite our best efforts, it is likely that our review is missing at least some relevant datasets.
We are committed to adding these datasets, along with future relevant dataset releases, to SafetyPrompts.com.%

\subsection{Recording Structured Information}
\label{subsec: methods - recording info}

For each of the 144 datasets in our review, we recorded 23 pieces of structured information.
At a high level, our goal was to capture the full development pipeline of each dataset: from how the dataset was created, to what entries in the dataset look like, what the dataset can or should be used for, how it can be accessed, and where it was published.
We show the full codebook in Appendix~\ref{app: code book}, which describes the structure and content of our main review spreadsheet.
We make the spreadsheet available along with code to reproduce our analyses at github.com/paul-rottger/safetyprompts-paper.

\section{Dataset Review Findings}
\label{sec: dataset review findings}

\begin{figure*}[ht]
    \includegraphics[width=0.9\textwidth]{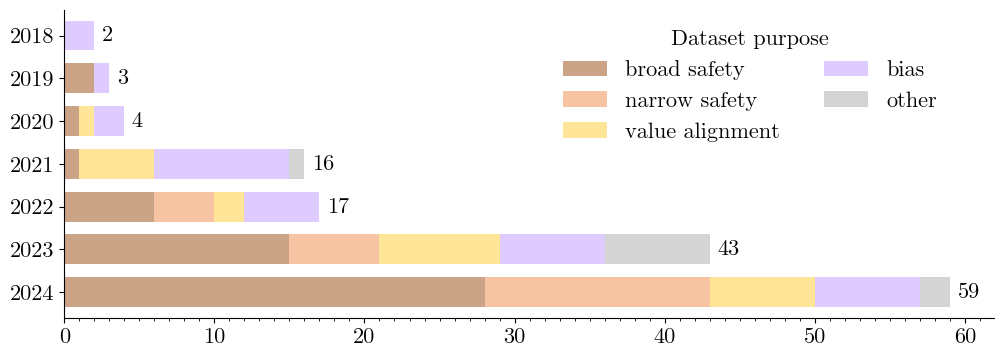}

    \vspace{0.3cm}
    \raggedleft
    
    \footnotesize
    \renewcommand{\arraystretch}{1.6}

    \begin{tabularx}{0.99\linewidth}{p{0.3cm}|X}

        \multirow{5}{*}{\rotatebox[origin=c]{90}{Example datasets}} & 
        \colorbox{lightbrown}{DecodingTrust} by \citet{wang2024decodingtrust} evaluates LLM safety across 8 ``perspectives''\\
        & \colorbox{lightorange}{ConfAIde} by \citet{mireshghallah2024confaide} evaluates the privacy-reasoning capabilities of LLMs\\
        & \colorbox{lightyellow}{MoralChoice} by \citet{scherrer2023evaluating} evaluates moral beliefs encoded in LLMs\\
        & \colorbox{lightpurple}{BBQ} by \citet{parrish2022bbq} evaluates social biases of LLMs in question answering\\
        & \colorbox{lightgray}{ToxicChat} by \citet{lin2023toxicchat} trains and evaluates content moderation systems for LLMs\\
        
    \end{tabularx}
    
    \caption{(top)
    \textbf{Number of datasets published per year, grouped by their primary purpose.}
    Our review includes 144 datasets published between June 2018 and December 2024.
    (bottom)
    \textbf{Example datasets} for each type of dataset purpose (\S\ref{subsec: findings - purpose}).
    }
    \label{fig:purpose_type}
\end{figure*}

\subsection{History and Growth of Safety Datasets}
\label{subsec: findings - timeline}

Our review shows that \textbf{LLM safety builds on a rich history of research into the risks and biases of language models and dialogue agents}.
The first datasets in our review were published in mid-2018, and focus on evaluating gender bias -- originally for co-reference resolution systems, but equally applicable to current LLMs \citep{zhao2018winobias,rudinger2018winogender}.
These datasets, in turn, build on earlier works on biases in word embeddings \citep[e.g.][]{bolukbasi2016man,caliskan2017semantics,garg2018word}, which illustrates that concerns around the negative social impacts of language models are far from new.
Even the term ``safety'' was already used by \citet{dinan2019build} and \citet{rashkin2019empatheticdialogues}, among others, who introduced datasets for evaluating and improving the safety of dialogue agents well before the current generative LLM paradigm.
By today's standards, however, interest in safety was relatively low at the time, with only 9 out of the 144 datasets in our review (6.3\%) published in or before 2020, as shown in Figure~\ref{fig:purpose_type}.

\textbf{LLM safety experienced a first moderate growth phase in 2021 and 2022}.
These two years, respectively, saw the publication of 16 and 17 open LLM safety datasets.
This coincides with increased interest in generative language models, particularly among researchers, following the release of GPT-3 in mid-2020 \citep{brown2020language}.

Right now, \textbf{research into LLM safety is experiencing unprecedented growth}.
This coincides with a surge in public interest in LLMs as well as concerns around LLM safety following the release of ChatGPT in November 2022.
43 out of the 144 datasets in our review (29.9\%) were published in 2023, and 2024 saw even more activity, with 59 datasets published.
Accordingly, it seems very likely that 2025 will surpass this record once more.

\subsection{Purpose of Datasets}
\label{subsec: findings - purpose}

In our review, we differentiate between five distinct dataset purposes:
\colorbox{lightbrown}{\textbf{Broad safety}} (n=53) denotes datasets that cover several aspects of LLM safety.
This includes structured evaluation datasets like SafetyKit \citep{dinan2022safetykit} or SimpleSafetyTests \citep{vidgen2024simplesafetytests} as well as broad-scope red-teaming datasets like BAD \citep{xu2021botadversarialdialogue} or AnthropicRedTeam \citep{ganguli2022red}.
\colorbox{lightorange}{\textbf{Narrow safety}} (n=25), conversely, denotes datasets that focus only on one specific aspect of LLM safety.
SafeText \citep{levy2022safetext}, for example, focuses only on commonsense physical safety, while SycophancyEval \citep{sharma2024sycophancyeval} focuses on sycophantic behaviour.
\colorbox{lightyellow}{\textbf{Value alignment}} (n=23) refers to datasets that are concerned with the ethical, moral or social behaviour of LLMs.
This includes datasets that seek to evaluate LLM understanding of ethical norms, like Scruples \citep{lourie2021scruples} and ETHICS \citep{hendrycks2021ethics}, as well as opinion surveys like GlobalOpinionQA \citep{durmus2024globalopinionqa}.
\colorbox{lightpurple}{\textbf{Bias}} (n=33) refers to datasets for evaluating sociodemographic biases in LLMs.
BOLD \citep{dhamala2021bold}, for example, evaluates bias in text completions, whereas DiscrimEval \citep{tamkin2023evaluating} evaluates biases in situated LLM decision-making.
\colorbox{lightgrey}{\textbf{Other}} (n=10), in our review, includes datasets for developing LLM chat moderation systems, like FairPrism \citep{fleisig2023fairprism} and ToxicChat \citep{lin2023toxicchat}, as well as specialised prompts from public competitions like Gandalf \citep{lakera2023gandalf}, Mosscap \citep{lakera2023mosscap} or HackAPrompt \citep{schulhoff2023hackaprompt}.

Figure~\ref{fig:purpose_type} shows that \textbf{early safety datasets were primarily concerned with evaluating biases}.
14 out of 25 datasets (56.0\%) published between 2018 and 2021 were created to identify and analyse sociodemographic biases in language models.
13 of these datasets evaluate gender biases, either exclusively \citep[e.g.][]{nozza2021honest} or along with other categories of bias such as race and sexual orientation \citep[e.g.][]{sheng2019woman} or religion \citep[e.g.][]{li2020unqovering}.

\textbf{Broad safety emerged as a prominent theme in 2022, driven by industry contributions}.
Anthropic, for instance, released two broad-scope red-teaming datasets \citep{ganguli2022red,bai2022training}, while Meta published datasets on positive LLM conversations \citep{ung2022saferdialogues} and general safety evaluation \citep{dinan2022safetykit}.
Most recently, broad safety has shifted towards more structured evaluation, as exemplified by benchmarks like DecodingTrust \citep{wang2024decodingtrust} or HarmBench \citep{mazeika2024harmbench}.

\textbf{There is now a trend towards more specialised safety evaluations}.
Narrow safety evaluations did not emerge until 2022, but now make up a significant portion of all new datasets. 
In 2024 alone, 15 of the 59 datasets in our review (25.4\%) were concerned with specific aspects of LLM safety, like rule-following \citep{mu2024rules} or privacy-reasoning ability \citep{mireshghallah2024confaide}.

Relatedly, \textbf{there is an increasing focus on adversarial robustness and LLM ``jailbreaking''}.
A growing body of work introduces safety datasets that are explicitly designed to elicit unsafe responses even from models trained to respond safely, often using elaborate prompting strategies, also known as ``jailbreaks'' \citep{wei2023jailbroken}.
Since August 2023, we record 13 such datasets, which differ from most other datasets that seek to evaluate safety in response to simpler, more naturalistic prompts \citep[e.g.][]{bhardwaj2023redteaming,shaikh2023harmfulq,vidgen2024simplesafetytests}.

\subsection{Intended Use of Datasets}
\label{subsec: findings - intended use}

\textbf{Most datasets are intended for model evaluation only}.
112 out of the 144 datasets in our review (77.8\%) were created for benchmarking or evaluation.
Only 6 datasets (4.2\%), by contrast, comprise examples of positive interactions between users and LLMs, created specifically for model training \citep{rashkin2019empatheticdialogues,ung2022saferdialogues,kim2022prosocialdialog,bianchi2024safetytuned,guo2024controllablepreferenceoptimization,jiang2024wildteaming}.
Overall, there is much more work on evaluating LLM safety than there is work on improving LLM safety.

\subsection{Dataset Format and Size}
\label{subsec: findings - format}

\textbf{The format of LLM safety datasets has changed alongside general trends in LLM development}.
Early datasets, especially those created for bias evaluation, often used autocomplete-style formats (n=9), where models are tasked to either fill a masked word or finish a sentence snippet \citep[e.g.][]{sheng2019woman,dhamala2021bold,kirk2021bias}.
Such formats, which are most compatible with early language models like BERT or GPT-2, have since mostly been replaced by chat-style prompts, like open-ended questions and instructions (n=88), as well as promptable multiple-choice questions (n=24), which better suit current generative LLMs.

Dataset size varies substantially across the 144 datasets in our review.
Smaller datasets ($\le$1k entries, n=42) tend to be narrow safety evaluation datasets, written by the dataset authors.
For example, the smallest dataset -- ControversialInstructions by \citet{bianchi2024safetytuned} -- comprises 40 author-written prompts to test the propensity of LLMs to generate hate speech.
Larger datasets ($\ge$100k entries, n=21), on the other hand, are often broader in scope, designed for training and evaluation, and created through templating or LLM generation (\S\ref{subsec: findings - creation}).
For example, the largest dataset -- WorldValuesBench by \citet{zhao2024worldvaluesbench} -- contains 21.5m multiple-choice prompts created programmatically from individual responses to the large-scale World Value Survey.

\subsection{Dataset Creation}
\label{subsec: findings - creation}

\textbf{The use of templates is a consistently popular method for generating safety evaluation data}.
26~out of the 144 datasets in our review (18.1\%) consist of human-written phrase or prompt templates, which are expanded through combination into larger evaluation datasets.
HolisticBias \citep{smith2022holisticbias}, for example, comprises 26 sentence templates (e.g.\ ``I am an [attribute1] who is [attribute2].''), which are combined with around 600 individual descriptor terms to create 459.8k test cases.
Generally, template approaches are most popular for bias evaluation -- 16 out of 33 bias evaluation datasets in our review use a template approach -- but recent work has also used templates for evaluating broad-scope LLM safety \citep{wang2024decodingtrust} or specific concerns such as privacy reasoning \citep{mireshghallah2024confaide}.

\textbf{A significant portion of recently-released datasets is fully synthetic}.
21 of the 102 datasets released since 2023 (20.6\%) consist of chat prompts or multiple-choice questions generated entirely by LLMs -- most commonly by some variant of GPT-3.5.
\citet{shaikh2023harmfulq}, for example, prompt GPT-3.5 to generate 200 harmful questions, which they use to explore safety in chain-of-thought question answering.
The resulting dataset resembles the human-written SimpleSafetyTests \citep{vidgen2024simplesafetytests}.
Generally, synthetic datasets vary in size and scope much like human-written datasets do.

Relatedly, instead of relying on static templates for data creation, \textbf{many recent datasets are more flexibly augmented using LLMs}.
\citet{bhatt2023purple}, for instance, expand a small expert-written set of cyberattack instructions into a larger set of 1,000 prompts using Llama-70b-chat \citep{touvron2023llama}.
\citet{wang2024decodingtrust} take a similar approach to build their large-scale DecodingTrust benchmark.

By contrast, \textbf{very few datasets comprise naturalistic user interactions with LLMs}.
Around half of the 144 datasets in our review are hand-written (n=70, 48.6\%), but mostly by the authors of the corresponding dataset publications.
A much smaller proportion of datasets is created by humans -- either crowdworkers \citep[e.g.][]{ganguli2022red,aroyo2023dices,kirk2024prism} or participants of online competitions \citep[e.g.][]{lakera2023gandalf,lakera2023mosscap} -- interacting more naturally with LLMs.

Finally, \textbf{there is a trend towards reusing existing datasets rather than creating original data}.
34~out of the 59 datasets published in 2024 (57.6\%) were entirely~(n=10) or partially~(n=24) sampled from older datasets.
QHarm by \citet{bianchi2024safetytuned}, for example, is taken entirely from AnthropicHarmlessBase \citep{bai2022training}. 
Another common approach is to sample existing data and augment it with templates \citep[e.g.][]{tedeschi2024alert} or LLM generations \citep[e.g.][]{yu2024cosafe}.
By contrast, just 10 out of 43 datasets published in 2023 (23.3\%) reused existing data.
This shift suggests a growing maturity of the LLM safety dataset landscape.

\begin{table*}[ht]
    
    \centering
    \footnotesize
    \renewcommand{\arraystretch}{1.4}

    \begin{minipage}{.5\linewidth}
        
        \begin{tabularx}{0.95\linewidth}{p{0.3cm}p{4.3cm}X}
                
            \hline
            \rowcolor[HTML]{decbff}
             & \textbf{Academic / Non-Profit Org.} & \textbf{n} \\
            \hline
            
            1 & Stanford University & 15\\
            \rowcolor[rgb]{0.95,0.95,0.95}
            1 & Allen Institute for AI & 15\\
            3 & UC Berkeley & 14\\
            \rowcolor[rgb]{0.95,0.95,0.95}
            4 & University of Washington & 13\\
            5 & Carnegie Mellon University & 12\\
            \bottomrule
            
        \end{tabularx}
    \end{minipage}%
    \begin{minipage}{.5\linewidth}
        \raggedleft
        \begin{tabularx}{0.95\linewidth}{p{0.3cm}p{4.3cm}X}
                
            \hline
            \rowcolor[HTML]{FCE597}
            & \textbf{Industry Org.} & \textbf{n} \\
            \hline
            
            1 & Meta* (prev. Facebook) & 13\\
            \rowcolor[rgb]{0.95,0.95,0.95}
            2 & Anthropic & 9\\
            3 & Microsoft* (incl. Research) & 7\\
            \rowcolor[rgb]{0.95,0.95,0.95}
            4 & Google* (incl. DeepMind) & 6\\
            5 & Alibaba & 4\\
            \bottomrule
            
        \end{tabularx}
    \end{minipage} 
    
    \caption{\textbf{Organisations that published the most open LLM safety datasets} among the 144 datasets in our review. 
    For each dataset, we count all affiliations for all co-authors.
    Stanford at n=15, for example, means that 15 datasets had a Stanford author.
    }
    \label{tab:prolific_orgs}
\end{table*}

\subsection{Dataset Languages}
\label{subsec: findings - language}

\textbf{The vast majority of open LLM safety datasets uses English language only}.
113 out of 144 datasets in our review (78.5\%) are exclusively in English.
10~datasets (6.9\%) focus only on Chinese \citep[e.g.][]{zhou2022cdial,xu2023cvalues,zhao2023chbias}, and one dataset each focuses only on Arabic, Swedish, Hindi, Korean, and French.
The 16 other datasets (11.1\%) cover English along with one or more other languages.
\citet{jin2024multilingualtrolleyproblems} cover 106 languages, although test cases are auto-translated from the English.
The other 143 datasets in our review together cover 31 languages.

\subsection{Data Licensing and Access}
\label{subsec: findings - access}

\textbf{When data is shared, usage licenses are mostly permissive}.
The most common license is the very permissive MIT License, which is used for 58 out of 144 datasets (40.3\%).
23 datasets (16.0\%) use the Apache 2.0 License, which provides additional patent protections.
43 datasets (29.9\%) use variants of a CC BY 4.0 License, which requires dataset users to provide appropriate credit and indicate if changes were made to the dataset.
Notably, 15 datasets (10.4\%) prohibit commercial usage with a CC BY-NC License.
Only 2 datasets (1.4\%) use a more restrictive custom license.
As of December 17th, 2024, 15 datasets (10.4\%) do not specify any license.%
\footnote{While conducting our review, we reached out to authors of all datasets that had not specified a license and encouraged them to add one. At least five authors added a license as a result.}

\textbf{GitHub is still the most popular platform for sharing LLM safety data}.
72 datasets (50.0\%) are available only on GitHub.
However, 55 datasets (38.2\%) are available on both GitHub and HuggingFace, and several recent datasets are shared on Hugging Face only \citep[e.g.][]{han2024wildguard,manerba2024social}, suggesting a shift in how data is shared.

\subsection{Dataset Publication Authors and Venues}
\label{subsec: findings - publication}

\textbf{Academic and non-profit organisations drive most of the creation of open LLM safety datasets}.
For 69 out of 144 datasets in our review (47.9\%), all authors of the corresponding publication were affiliated only with academic or non-profit organisations.
45 datasets (31.3\%) included authors from industry and academia, and only 30 datasets (20.8\%) were published by fully industry teams.

\textbf{The creation of LLM safety datasets is concentrated in few research hubs} (Table~\ref{tab:prolific_orgs}).
There are 156 unique affiliations across the authors of the 144 datasets in our review.
101 affiliations (64.7\%) are associated with just a single dataset.
The five most prolific organisations, on the other hand, are each associated with at least 13 datasets.
All of the twenty most prolific organisations are located and/or headquartered in the US, with the exception of Bocconi University (Italy, n=11), the University of Oxford (UK, n=5), CUHK (HK, n=4), and Tsinghua University (China, n=4).

\textbf{Most LLM safety datasets so far have been published at *ACL conferences}.
68 out of the 144 datasets in our review (47.2\%) were published at either ACL (n=27), EMNLP (n=30) or other *ACL venues.
33 datasets (19.7\%) were published at more ML-focused conferences, i.e.\ NeurIPS (n=18), ICLR (n=10), or ICML (n=5).
Only 11 datasets (7.6\%) were published at other venues, and only 2 datasets appeared in journal publications \citep{jin2024kobbq,yu2024beyond}.
28 datasets (19.4\%), on the other hand, were accompanied only by arXiv preprints, and 4 (2.7\%) only by blog posts, meaning they did not receive traditional peer review.
Generally, we observe a slight trend away from *ACL conferences and towards more ML-focused venues as well as arXiv-only publication, although this could in part be explained by recent arXiv preprints still being under review.

\begin{table*}[ht]
    
    \centering
    \footnotesize
    \renewcommand{\arraystretch}{1.4}

    \begin{tabularx}{\linewidth}{p{0.4cm}p{5.6cm}Xp{0.4cm}}
            
        \hline
        \rowcolor[HTML]{f6c4a2}
         & \textbf{Dataset} & \textbf{Purpose} & \textbf{n} \\
        \hline
        1 & TruthfulQA \citep{lin2022truthfulqa} & evaluate tendency to mimic human falsehoods & 8 \\
        \rowcolor[rgb]{0.95,0.95,0.95}
        2 & BBQ \citep{parrish2022bbq} & evaluate social bias in question answering & 6 \\
        3 & AnthropicRedTeam \citep{ganguli2022red} & evaluate responses to diverse red-team attacks & 3 \\
        \rowcolor[rgb]{0.95,0.95,0.95}
        4 & ToxiGen \citep{hartvigsen2022toxigen} & evaluate toxicity in text completions & 3 \\
        5 & RealToxicityPrompts \citep{gehman2020realtoxicityprompts} & evaluate toxicity in text completions & 3 \\
        \bottomrule
        
    \end{tabularx}
    
    \caption{\textbf{Most popular open LLM safety datasets}, based on how often
    release publications for SOTA LLMs reported results on each dataset.
    BBQ at n=6, for example, means that 6 out of the 29 model release publications (\S\ref{subsec: datasets in model releases - findings}) reported BBQ results.
    }
    \label{tab:model_top_datasets}
\end{table*}

\section{Datasets in Model Release Publications}
\label{sec: datasets in model releases}

In the following, we briefly examine \textbf{how open LLM safety datasets are used in practice}.
In particular, we examine in this section which safety datasets are used to evaluate current state-of-the-art LLMs ahead of their release, as documented in model release publications.
In the next section (\S\ref{sec: datasets in model benchmarks}), we then examine which safety datasets are included in popular LLM benchmarks and leaderboards.
This is to characterise current norms and common practices in evaluating LLM safety, so that we can then discuss (\S\ref{sec: discussion}) these norms and practices in relation to the findings of our dataset review (\S\ref{sec: dataset review findings}).

\subsection{Scope of our Model Release Review}
\label{subsec: datasets in model releases - scope}

We examine the top 50 best-performing LLMs listed on the LMSYS Chatbot Arena Leaderboard \citep{chiang2024chatbotarena} as of July 25th, 2024.%
\footnote{\url{https://lmarena.ai/?leaderboard}}
The LMSYS Chatbot Arena Leaderboard is a crowdsourced platform for LLM evaluation, which ranks models based on model Elo scores calculated from over one million pairwise human preference votes.
We use this leaderboard for setting the scope of our review because it is held in high regard in the LLM community, and it has up-to-date coverage of recent model releases.

The top 50 entries on the LMSYS leaderboard correspond to 29 unique model releases.%
\footnote{A model release may comprise multiple model versions, such as GPT-4-0314 and GPT-4-0613.}
Of these 29 models, 16 (55.2\%) are proprietary models only accessible via an API, released by companies such as OpenAI (GPT), Anthropic (Claude), Google (Gemini) and 01 AI (Yi).
The other 13 models (44.8\%) are open models, for which weights are publicly accessible  via Hugging Face.
Proprietary models generally outrank open models on the leaderboard, with Gemma 2 27b \citep{deepmind2024gemma2} being the best open model at rank 14.
All 29 models were released by industry labs.%

\subsection{Findings of our Model Release Review}
\label{subsec: datasets in model releases - findings}

\textbf{Around half of state-of-the-art LLMs are evaluated for safety ahead of their release}, with substantial variation in the extent and nature of safety evaluations.
For 16 out of 29 model releases (55.2\%), model developers report at least some quantitative safety evaluation in the associated technical report, blog post or model card.
13 model release publications (44.8\%) report results on at least one open LLM safety dataset.
The Claude 3.5 model card, for example, reports results on WildChat \citep{zhao2024inthewildchat} and XSTest \citep{rottger2024xstest}, whereas Gemma 2 \citep{deepmind2024gemma2} is evaluated on 7 different open LLM safety datasets.
13 out of the 29 models (44.8\%), on the other hand, do not report any safety evaluations.
This includes the strongest model, the proprietary GPT-4o, which mentions safety measures in its release blog post, but lacks any concrete quantitative safety evaluations.
Other models, both proprietary (e.g.\ Yi-Large, Reka Core) and open (e.g.\ Mixtral, Command R), do not mention safety at all in their release publications.

When safety is evaluated, \textbf{proprietary data plays a large role in model release safety evaluations}.
Out of the 16 model releases that report safety evaluation results, 11 (68.8\%) use undisclosed proprietary data for evaluating model safety.
3 of these releases -- Llama 3 \citep{meta2024llama3}, Qwen 2 \citep{yang2024qwen2}, and Phi 3 \citep{abdin2024phi3} -- report results only on proprietary safety datasets.

Finally, \textbf{the diversity of open LLM safety datasets used in model release evaluations is very limited}. 
Only 14 open LLM safety datasets are used across the 29 model releases, and 6 of these 14 datasets are used only once.
Table~\ref{tab:model_top_datasets} shows the 5 datasets that are used most often.
Notably, TruthfulQA \citep{lin2022truthfulqa} is used in 8 out of the 16 model release publications that report any safety evaluation results (50.0\%), and it is often framed by model developers as a capability rather than a safety evaluation.
We discuss the implications of these findings in \S\ref{sec: discussion}.

\section{Datasets used in Popular Benchmarks}
\label{sec: datasets in model benchmarks}

\subsection{Scope of our Benchmark Review}
\label{subsec: datasets in model benchmarks - scope}

To complement our model release review, we briefly examine 9 popular LLM benchmarks for which safety datasets they include.
6 benchmarks are widely-used general-purpose benchmarking suites: Stanford's HELM Classic \citep{liang2023helm} and HELM Instruct \citep{zhang2024helminstruct}, Hugging Face's Open LLM Leaderboard \citep{beeching2023openllmleaderboard}, AllenAI's RewardBench \citep{lambert2024rewardbench}, Eleuther AI's Evaluation Harness \citep{gao2021evalharness}, and BIG-Bench \citep{srivastava2023bigbench}.
3 benchmarks focus explicitly on LLM safety: TrustLLM \citep{sun2024trustllm}, HELM Safety \citep{kaiyom2024helmsafety}, and the LLM Safety Leaderboard.\footnote{\url{https://huggingface.co/spaces/AI-Secure/llm-trustworthy-leaderboard}, based on \citet{wang2024decodingtrust}.}

\subsection{Findings of our Benchmark Review}
\label{subsec: datasets in model benchmarks - findings}

\textbf{There are large differences in how different benchmarks evaluate LLM safety}.
The 9 benchmarks make use of 26 open LLM safety datasets.
17 of these datasets are used in just one benchmark.
TrustLLM \citep{sun2024trustllm}, for example, combines 11 open LLM safety datasets, of which 5 are not used in any other benchmark.
The only open LLM safety datasets that are used in more than 2 benchmarks are TruthfulQA \citep{lin2022truthfulqa}, which is used in 4 benchmarks, as well as RealToxicityPrompts \citep{gehman2020realtoxicityprompts}, BBQ \citep{parrish2022bbq}, ETHICS \citep{hendrycks2021ethics}, XSTest \citep{rottger2024xstest}, and DoNotAnswer \citep{wang2024donotanswer}, which are each used in 3 benchmarks.

\textbf{There is currently no single LLM safety benchmark with a truly comprehensive scope}.
The TrustLLM benchmark \citep{sun2024trustllm}, has the broadest scope relevant to safety among the 9 benchmarks we examined, covering malicious instruction-following, bias, and value alignment.
However, it does not, for example, test for longer-term risk potential with evaluations for sycophancy \citep{perez2023discovering} or chemical weapon knowledge \citep{li2024wmdp}, as the Evaluation Harness does \citep{gao2021evalharness}.
Similarly, the LLM Safety Leaderboard \citep{beeching2023openllmleaderboard} tests for toxic content generation and malicious instruction-following, but not for false refusal \citep[e.g.][]{rottger2024xstest} or sociodemographic biases \citep[e.g.][]{parrish2022bbq}

\section{Discussion \& Future Directions}
\label{sec: discussion}

\subsection{The State of the Safety Dataset Landscape}
\label{subsec: discussion - datasets}

Overall, our review shows that \textbf{growing interest in LLM safety is driving the creation of more and more diverse open LLM safety datasets}.
More datasets were published in 2023 than ever before, and 2024 again surpassed this record (\S\ref{subsec: findings - timeline}).
Existing datasets span varied purposes (\S\ref{subsec: findings - purpose}) and formats (\S\ref{subsec: findings - format}), which have adapted over time to meet the needs and requirements of LLM users and developers.
Researchers and practitioners are making creative use of new methods for dataset creation (\S\ref{subsec: findings - creation}), and when data is shared, usage licenses are mostly permissive (\S\ref{subsec: findings - access}).
These are encouraging signs for the health of the open LLM safety community and its ability to address emerging challenges and fill gaps in dataset coverage as they become apparent.

Among these gaps, the most apparent today is that \textbf{there is a clear lack of safety datasets in non-English languages}.
We found that English dominates the current safety dataset landscape (\S\ref{subsec: findings - language}), mirroring long-standing trends in NLP research \citep{bender2011achieving,joshi2020state,holtermann2024evaluating}.
To some extent, this imbalance also reflects an imbalance in who is publishing safety datasets (\S\ref{subsec: findings - publication}).
The lack of non-English resources for evaluating and improving LLM safety is a problem because it means that billions of non-English speakers across the world are potentially more at risk of harm when using current language technologies.
Further, interpretations of safety are known to vary across cultures and geographies \citep{aroyo2023dices,kirk2024prism}.
These two factors create an urgent need for the LLM safety community to prioritise the creation of non-English datasets going forward.
Recent datasets like AyaRedTeaming \citep{aakanksha2024multilingual}, which were created for language-specific cultural contexts with the involvement of local stakeholders, may serve as a useful blueprint for future work in this direction.

The second major concern apparent from our review is that \textbf{current safety evaluation datasets largely fail to reflect real-world LLM usage}, which undermines their ecological validity.
Data generation and augmentation, from templates to more recent LLM-based methods, have enabled rapid dataset creation (\S\ref{subsec: findings - creation}), but it is not clear that they can emulate the ``messiness'' and diversity that is apparent in real-world user interactions with LLMs \citep{ouyang2023shifted,zhao2024inthewildchat,zheng2024lmsys}.
An increasing focus on safety against ``jailbreaking'' (\S\ref{subsec: findings - purpose}) may match the behaviour of sophisticated adversarial users, but does not capture the potential risks from vulnerable populations interacting with LLMs without malicious intent \citep{vidgen2024mlc05}.
Addressing these concerns will require the creation of more naturalistic evaluations, and clearer communication about which user personas (e.g.\ vulnerable, malicious, or adversarial) are targeted by individual safety evaluations. 
Encouraging developments in this direction include WildGuard \citep{han2024wildguard} and WildJailBreak \citep{jiang2024wildteaming}, which provide evaluations based on real user prompts. 
Future work may similarly look to collections of real user interactions with LLMs, such as WildChat \citep{zhao2024inthewildchat} and LMSYS \citep{zheng2024lmsys}, to improve the ecological validity of LLM safety evaluations.

\subsection{The Use of Safety Datasets in Practice}
\label{subsec: discussion - benchmarks}

Our analysis of how open LLM safety datasets are used in practice shows that \textbf{there is clear scope for standardisation in LLM safety evaluations}.
Safety is a key priority for model developers and users, as evidenced by the inclusion of safety evaluations in the majority of model release publications (\S\ref{sec: datasets in model releases}) and popular LLM benchmarks (\S\ref{sec: datasets in model benchmarks}).
However, there is much idiosyncrasy in \textit{which} datasets are used for evaluating safety across model release publications and benchmarks.
For commercial model releases, these datasets are often proprietary and undisclosed.
More standardised and open evaluations, on the other hand, would enable more meaningful model comparisons and incentivise the development of safer LLMs.

We see three main directions towards this goal.
First, current evaluation practices could \textbf{better leverage recent progress in safety dataset creation}.
The most popular open datasets for evaluating safety in model release publications, for example, are all from 2020 or 2022 (Table~\ref{tab:model_top_datasets}), despite more than two thirds of the datasets in our review being published in or after 2023.
Prior publication date is not a sign of lacking quality, but older autocomplete-style datasets like RealToxicityPrompts \citep{gehman2020realtoxicityprompts} or ToxiGen \citep{hartvigsen2022toxigen} simply do not reflect realistic usage of current chat-optimised LLMs \citep{ouyang2023shifted,zheng2024lmsys,zhao2024inthewildchat}, which undermines their ecological validity.
Our review highlights many datasets that could be used in their stead.
Second, there is a clear need for \textbf{unified standards of dataset quality}.
In this paper, we refrained from making quality judgments about the datasets we reviewed, mainly because different datasets serve different purposes, so that their utility is highly context dependent.
However, recent efforts like BetterBench \citep{reuel2024betterbench} as applied to safety datasets could help create consensus on which datasets to use for safety benchmarking.
Third, safety research would benefit from \textbf{unified safety evaluation protocols}.
The open-ended nature of safety-relevant behaviours complicates standardised evaluation compared to many factual tasks.
Frameworks like Eleuther AI's Evaluation Harness, adapted for safety tasks, would make it easier for model developers and users to run the same safety evaluations with reproducible results, and thus serve to reduce idiosyncrasy in prevailing safety evaluation practices.

\section{Conclusion}
\label{sec: conclusion}

In recent years, researchers and practitioners have sought to meet concerns around the safety of large language models by creating an abundance of datasets for evaluating and improving LLM safety.
However, the rapid pace of dataset creation and the variety of purposes served by different datasets have made it difficult for researchers and practitioners to find the most relevant datasets for different use cases, and to identify gaps in dataset coverage that future work may fill.
In this paper, we addressed these issues by conducting a first systematic review of open LLM safety datasets.

In our review, which includes 144 datasets published between June 2018 and December 2024, we showed that, encouragingly, existing datasets span varied purposes and formats, which have adapted over time to meet the changing needs and requirements of LLM users and developers.
However, we also highlighted major outstanding challenges, including a clear lack of non-English datasets as well as naturalistic safety evaluations.
Further, when examining how open LLM safety datasets are used in practice -- in model release publications and popular LLM benchmarks -- we found that current evaluation practices are highly idiosyncratic and make use of only a small fraction of available datasets, which presents clear scope for rejuvenation and standardisation.

Overall, we hope that with our review, as well as the living dataset catalogue we make available on SafetyPrompts.com, we can enable such positive change, by helping researchers and practitioners make the best use of existing datasets, and providing a strong foundation for future dataset development.

\section*{Acknowledgments}

Thank you for feedback and dataset suggestions to
Giuseppe Attanasio, Steven Basart, Federico Bianchi, Marta R. Costa-Jussà, Daniel Hershcovic, Kexin Huang, Hyunwoo Kim, George Kour, Bo Li, Hannah Lucas, Marta Marchiori Manerba, Norman Mu, Niloofar Mireshghallah, Matus Pikuliak, Verena Rieser, Felix Röttger, Sam Toyer, Ryan Tsang, Pranav Venkit, Laura Weidinger, and Linhao Yu.
Special thanks to Hannah Rose Kirk for the initial logo suggestion.

PR, FP, and DH are members of the Data and Marketing Insights research unit of the Bocconi Institute for Data Science and Analysis, and are supported by a MUR FARE 2020 initiative under grant agreement Prot.\ R20YSMBZ8S (INDOMITA) and the European Research Council (ERC) under the European Union’s Horizon 2020 research and innovation program (No.\ 949944, INTEGRATOR).

\bibliography{custom}

\appendix

\section{Review Code Book}
\label{app: code book}

For each of the 144 datasets in our review, we recorded 23 pieces of structured information.
At a high level, our goal was to capture the full development pipeline of each dataset: from how the dataset was created, to what entries in the dataset look like, what the dataset can or should be used for, how it can be accessed, and where it was published.
We show the full codebook in Table~\ref{tab:code_book} below, which describes the structure and content of our main review spreadsheet.
We make the spreadsheet available along with code to reproduce our analyses at github.com/paul-rottger/safetyprompts-paper.

\begin{table*}[ht]

    \centering
    \footnotesize
    \renewcommand{\arraystretch}{1.4}
    
    \begin{tabularx}{\textwidth}{p{1.9cm}p{6.5cm}X}
        
        \hline
        \rowcolor[HTML]{FCE597}
        \multicolumn{3}{c}{\textbf{\texttt{purpose}}: what can I use this dataset for?} \\
        \hline
        
        \texttt{\_type} & High-level type of purpose / \newline general area of application & single choice: [broad safety, narrow safety, bias, value alignment, other] \\
        \rowcolor[rgb]{0.95,0.95,0.95}
        \texttt{\_tags} & Additional tags to specify purpose & comma-separated list of text tags \\
        \texttt{\_stated} & Exact purpose as stated by authors & free text \\
        \rowcolor[rgb]{0.95,0.95,0.95}
        \texttt{\_llmdev} & Intended use of the dataset within the LLM development pipeline &  single choice: [eval only, train and eval, train only, other] \\

        \hline
        \rowcolor[HTML]{FCE597}
        \multicolumn{3}{c}{\textbf{\texttt{entries}}: what do entries in this dataset look like?} \\
        \hline

        \texttt{\_type} & High-level type of entry / format & single choice: [chat, multiple choice, autocomplete, other] \\
        \rowcolor[rgb]{0.95,0.95,0.95}
        \texttt{\_languages} & Languages in the dataset & comma-separated list of free-text language names \\
        \texttt{\_n} & Number of entries & integer \\
        \rowcolor[rgb]{0.95,0.95,0.95}
        \texttt{\_unit} & Unit of entries (e.g.\ conversation) & free text \\
        \texttt{\_detail} & Additional detail on entry format & free text \\

         \hline
        \rowcolor[HTML]{FCE597}
        \multicolumn{3}{c}{\textbf{\texttt{creation}}: who created this dataset / where is it sampled from?} \\
        \hline

        \texttt{\_creator\_type} & Type of creator, i.e.\ who or what created the data & single choice: [human, machine, hybrid] \\
        \rowcolor[rgb]{0.95,0.95,0.95}
        \texttt{\_source\_type} & Type of data source, i.e.\ where the data is taken from & single choice: [original, sampled, mixed] \\
        \texttt{\_detail} & Additional detail on creator/source & free text \\

        \hline
        \rowcolor[HTML]{FCE597}
        \multicolumn{3}{c}{\textbf{\texttt{access}}: where can I download this dataset, and how is it licensed?} \\
        \hline

        \texttt{\_git\_url} & GitHub repo URL & URL \\
        \rowcolor[rgb]{0.95,0.95,0.95}
        \texttt{\_hf\_url} & Hugging Face dataset URL & URL \\
        \texttt{\_license} & Dataset license & free text \\

        \hline
        \rowcolor[HTML]{FCE597}
        \multicolumn{3}{c}{\textbf{\texttt{publication}}: when, where, and by whom was this dataset published?} \\
        \hline

        \texttt{\_date} & Publication date (most recent version) & dd-mmm-yyyy \\
        \rowcolor[rgb]{0.95,0.95,0.95}
        \texttt{\_affils} & Author affiliations & comma-separated list of institutions \\
        \texttt{\_sector} & Sector from which the publication originated & single choice: [academia, industry, mixed] \\
        \rowcolor[rgb]{0.95,0.95,0.95}
        \texttt{\_name} & Publication name / reference & free text \\
        \texttt{\_venue} & Publication venue & free text \\
        \rowcolor[rgb]{0.95,0.95,0.95}
        \texttt{\_url} & Publication URL & URL \\

        \hline
        \rowcolor[HTML]{FCE597}
        \multicolumn{3}{c}{\textbf{\texttt{other}}: what is worth noting beyond the scope of this review?} \\
        \hline

        \texttt{\_notes} & Additional notes & free text \\
        \rowcolor[rgb]{0.95,0.95,0.95}
        \texttt{\_date\_added} & Date on which the dataset was added to SafetyPrompts.com & dd-mmm-yyyy \\
        
        \bottomrule
    \end{tabularx}
    \caption{\textbf{Codebook for our main review spreadsheet}.
    For each of the 144 datasets included in our review, we recorded 23 pieces of structured information.}
    \label{tab:code_book}
\end{table*}

\section{List of All Datasets}
\label{app: datasets}

In total, our review covers 144 datasets published before our cutoff date of December 17th, 2024.
We did not cite all dataset papers in the main body of our review, so we cite all datasets here, in ascending order by year of publication.

\begin{enumerate}
    \item JBBBehaviours \citep{chao2024jailbreakbench}
    \item MedSafetyBench \citep{han2024medsafetybench}
    \item PRISM \citep{kirk2024prism}
    \item WildGuardMix \citep{han2024wildguard}
    \item SGBench \citep{mou2024sgbench}
    \item StrongREJECT \citep{souly2024strongreject}
    \item CoCoNot \citep{brahman2024coconot}
    \item WildJailBreak \citep{jiang2024wildteaming}
    \item MultiTP \citep{jin2024multilingualtrolleyproblems}
    \item DoAnythingNow \citep{shen2024dan}
    \item ForbiddenQuestions \citep{shen2024dan}
    \item CoSafe \citep{yu2024cosafe}
    \item SoFa \citep{manerba2024social}
    \item ArabicAdvBench \citep{ghanim2024arabicadvbench}
    \item AyaRedTeaming \citep{aakanksha2024multilingual}
    \item UltraSafety \citep{guo2024controllablepreferenceoptimization}
    \item GEST \citep{pikuliak2024gest}
    \item DeMET \citep{levy2024demet}
    \item GenMO \citep{bajaj2024genmo}
    \item SGXSTest \citep{gupta2024walledeval}
    \item HindiXSTest \citep{gupta2024walledeval}
    \item CIVICS \citep{pistilli2024civics}
    \item GlobalOpinionQA \citep{durmus2024globalopinionqa}
    \item CALM \citep{gupta2024calm}
    \item ChiSafetyBench \citep{zhang2024chisafetybench}
    \item SafetyBench \citep{zhang2024safetybench}
    \item CatQA \citep{bhardwaj2024catqa}
    \item KorNAT \citep{lee2024kornat}
    \item CMoralEval \citep{yu2024cmoraleval}
    \item XSafety \citep{wang2024xsafety}
    \item SaladBench \citep{li2024saladbench}
    \item MMHB \citep{tan2024mmhb}
    \item GPTFuzzer \citep{yu2024gptfuzzer}
    \item WMDP \citep{li2024wmdp}
    \item ALERT \citep{tedeschi2024alert}
    \item ORBench \citep{cui2024orbench}
    \item SorryBench \citep{xie2024sorrybench}
    \item XSTest \citep{rottger2024xstest}
    \item Flames \citep{huang2024flames}
    \item SEval \citep{yuan2024seval}
    \item WorldValuesBench \citep{zhao2024worldvaluesbench}
    \item CBBQ \citep{huang2024cbbq}
    \item KoBBQ \citep{jin2024kobbq}
    \item SAFE \citep{yu2024safe}
    \item AegisAIContentSafety \citep{ghosh2024aegis}
    \item DoNotAnswer \citep{wang2024donotanswer}
    \item RuLES \citep{mu2024rules}
    \item HarmBench \citep{mazeika2024harmbench}
    \item DecodingTrust \citep{wang2024decodingtrust}
    \item SimpleSafetyTests \citep{vidgen2024simplesafetytests}
    \item CoNA \citep{bianchi2024safetytuned}
    \item ControversialInstructions \citep{bianchi2024safetytuned}
    \item QHarm \citep{bianchi2024safetytuned}
    \item MaliciousInstructions \citep{bianchi2024safetytuned}
    \item PhysicalSafetyInstructions \citep{bianchi2024safetytuned}
    \item SafetyInstructions \citep{bianchi2024safetytuned}
    \item HExPHI \citep{qi2024hexphi}
    \item MaliciousInstruct \citep{huang2024maliciousinstruct}
    \item SycophancyEval \citep{sharma2024sycophancyeval}
    \item ConfAIde \citep{mireshghallah2024confaide}
    \item OKTest \citep{shi2024oktest}
    \item AdvBench \citep{zou2023advbench}
    \item Mosscap \citep{lakera2023mosscap}
    \item TDCRedTeaming \citep{mazeika2023tdc}
    \item JADE \citep{zhang2023jade}
    \item CPAD \citep{liu2023cpad}
    \item CyberattackAssistance \citep{bhatt2023purple}
    \item AttaQ \citep{kour2023attaq}
    \item AART \citep{radharapu2023aart}
    \item DELPHI \citep{sun2023delphi}
    \item AdvPromptSet \citep{esiobu2023robbie}
    \item HolisticBiasR \citep{esiobu2023robbie}
    \item DiscrimEval \citep{tamkin2023discrimeval}
    \item HackAPrompt \citep{schulhoff2023hackaprompt}
    \item ToxicChat \citep{lin2023toxicchat}
    \item SPMisconceptions \citep{chen2023spmisconceptions}
    \item FFT \citep{cui2023fft}
    \item MoralChoice \citep{scherrer2023evaluating}
    \item DICES350 \citep{aroyo2023dices}
    \item DICES990 \citep{aroyo2023dices}
    \item BeaverTails \citep{ji2023beavertails}
    \item PromptExtractionRobustness \citep{toyer2023tensor}
    \item PromptHijackingRobustness \citep{toyer2023tensor}
    \item GandalfIgnoreInstructions \citep{lakera2023gandalf}
    \item GandalfSummarization \citep{lakera2023gandalf}
    \item HarmfulQA \citep{bhardwaj2023redteaming}
    \item LatentJailbreak \citep{qiu2023latentjailbreak}
    \item OpinionQA \citep{santurkar2023opinionqa}
    \item CValuesResponsibilityMC \citep{xu2023cvalues}
    \item CValuesResponsibilityPrompts \citep{xu2023cvalues}
    \item CHBias \citep{zhao2023chbias}
    \item HarmfulQ \citep{shaikh2023harmfulq}
    \item FairPrism \citep{fleisig2023fairprism}
    \item ModelWrittenSycophancy \citep{perez2023discovering}
    \item ModelWrittenAdvancedAIRisk \citep{perez2023discovering}
    \item ModelWrittenPersona \citep{perez2023discovering}
    \item WinoGenerated \citep{perez2023discovering}
    \item SeeGULL \citep{jha2023seegull}
    \item WinoQueer \citep{felkner2023winoqueer}
    \item Machiavelli \citep{pan2023machiavelli}
    \item SafetyPrompts \citep{sun2023safetyprompts}
    \item OIGModeration \citep{ontocord2023oigmoderation}
    \item CDialBias \citep{zhou2022cdial}
    \item HolisticBias \citep{smith2022holisticbias}
    \item PersonalInfoLeak \citep{huang2022personalinfoleak}
    \item ProsocialDialog \citep{kim2022prosocialdialog}
    \item SafeText \citep{levy2022safetext}
    \item AnthropicRedTeam \citep{ganguli2022red}
    \item IndianStereotypes \citep{bhatt2022indianstereotypes}
    \item MoralExceptQA \citep{jin2022moralexceptqa}
    \item BBQ \citep{parrish2022bbq}
    \item FrenchCrowSPairs \citep{neveol2022frenchcrows}
    \item MIC \citep{ziems2022moralintegrity}
    \item SaFeRDialogues \citep{ung2022saferdialogues}
    \item SafetyKit \citep{dinan2022safetykit}
    \item ToxiGen \citep{hartvigsen2022toxigen}
    \item DiaSafety \citep{sun2022diasafety}
    \item TruthfulQA \citep{lin2022truthfulqa}
    \item AnthropicHarmlessBase \citep{bai2022training}
    \item JiminyCricket \citep{hendrycks2022jiminy}
    \item BiasOutOfTheBox \citep{kirk2021bias}
    \item EthnicBias \citep{ahn2021ethnicbias}
    \item ConvAbuse \citep{cercas2021convabuse}
    \item MoralStories \citep{emelin2021moralstories}
    \item RedditBias \citep{barikeri2021redditbias}
    \item HypothesisStereotypes \citep{sotnikova2021hypothesisstereotypes}
    \item StereoSet \citep{nadeem2021stereoset}
    \item ETHICS \citep{hendrycks2021ethics}
    \item BAD \citep{xu2021botadversarialdialogue}
    \item HONEST \citep{nozza2021honest}
    \item SweWinoGender \citep{hansson2021swewinogender}
    \item LMBias \citep{liang2021lmbias}
    \item ScruplesAnecdotes \citep{lourie2021scruples}
    \item ScruplesDilemmas \citep{lourie2021scruples}
    \item BOLD \citep{dhamala2021bold}
    \item CrowSPairs \citep{nangia2020crows}
    \item UnQover \citep{li2020unqovering}
    \item RealToxicityPrompts \citep{gehman2020realtoxicityprompts}
    \item SocialChemistry101 \citep{forbes2020socialchemistry}
    \item Regard \citep{sheng2019woman}
    \item ParlAIDialogueSafety \citep{dinan2019build}
    \item EmpatheticDialogues \citep{rashkin2019empatheticdialogues}
    \item WinoBias \citep{zhao2018winobias}
    \item WinoGender \citep{rudinger2018winogender}
    
\end{enumerate}

\end{document}